\documentclass[11pt,a4paper]{article}
\usepackage[hyperref]{eacl2021}
\usepackage{times}
\usepackage{latexsym}
\usepackage[linesnumbered,ruled,vlined]{algorithm2e}
\usepackage{caption}
\usepackage{adjustbox,lipsum}

\usepackage{amsmath}
\allowdisplaybreaks

\usepackage{microtype}

\aclfinalcopy 

\title{Dual-State Capsule Networks for Text Classification}

\author{Piyumal Demotte, Surangika Ranathunga \\
  Department of Computer Science and Engineering, University of Moratuwa \\
  Katubedda 10400, Sri Lanka \\
  \texttt{\{piyumalanthony.16,surangika}\}@cse.mrt.ac.lk}

\date{}

\begin{document}
\maketitle
\begin{abstract}
Text classification systems based on contextual embeddings are not viable options for many of the low resource languages. On the other hand, recently introduced capsule networks have shown performance in par with these text classification models. Thus, they could be considered as a viable alternative for text classification for languages that do not have pre-trained contextual embedding models. However, current capsule networks depend upon spatial patterns without considering the sequential features of the text. They are also sub-optimal in capturing the context-level information in longer sequences. This paper presents a novel Dual-State Capsule (DS-Caps) network-based technique for text classification, which is optimized to mitigate these issues. Two varieties of states, namely sentence-level and word-level, are integrated with capsule layers to capture deeper context-level information for language modeling. The dynamic routing process among capsules was also optimized using the context-level information obtained through sentence-level states. The DS-Caps networks outperform the existing capsule network architectures for multiple datasets, particularly for tasks with longer sequences of text. We also demonstrate the superiority of DS-Caps in text classification for a low resource language. 
\end{abstract}

\section{Introduction}\label{intro}

Recent research has shown state-of-the-art results using BERT-like contextual embedding models~\cite{devlin2018bert}. However, these performances highly depend on the comprehensiveness of the pre-trained contextual embedding models. In other words, these models require a very large corpus to produce optimal results in down-stream tasks such as text classification. However, for many low resource languages, such large corpora are not available. Even if available, building contextual embedding models for each and every language has practical concerns with respect to resource requirements. As a solution, multilingual models\footnote{\url{https://github.com/google-research/bert/blob/master/multilingual.md}} were introduced. However, still there are many languages that are not included in these publicly available multilingual models. It has also been shown that these multilingual models give sub-optimal results compared to those trained on monolingual data~\cite{dumitrescu2020birth}. Thus, text classification for these languages still has to rely on techniques that do not involve BERT-like models.

Another line of research has used language-specific linguistic features such as Part of Speech (POS) as auxiliary input to neural models~\cite{qian2016linguistically}. Although these have also shown very impressive results, low resource languages that do not have such resources cannot take benefit of this line of research as well.

Among the Deep Learning models for text classification that do not rely on contextual embeddings or auxiliary linguistic features, capsule networks are in the fore-front~\cite{zhao2018investigating, kim2020text}. For example, for the MR(2005)~\cite{pang2005seeing} results produced by Capsule Network models closely trail behind linguistically enhanced and contextual embeddings-based Deep Learning techniques~\cite{zhao2018investigating}.


This paper further improves the state-of-the-art capsule networks and presents a novel capsule network architecture namely, Dual-State Capsule Networks (DS-Caps). In particular, this is an enhancement to the existing capsule networks model with dynamic routing, which has been employed for text classification~\cite{zhao2018investigating}. DS-Caps is mainly inspired by the Sentence State LSTM (S-LSTM) networks~\cite{zhang2018sentence}, which eliminates the sequential dependencies of LSTMs to enable to capturing local n-grams as well as sequential features of the text simultaneously. The proposed model could be represented as a sophisticated language model that elevates the usage of n-grams, sequential patterns, and capsules vector representation for language modeling within a single architecture. DS-Caps further mitigate the limitations with the vanilla capsule network of capturing contextual-level data of the text while sequential processing.

We empirically show that our approach produces state-of-the-art performances against the existing capsule networks, for many publicly available datasets for text classification tasks. Furthermore, DS-Caps is evaluated against a text classification task for a low resource language, for which there are no pre-trained monolingual or multilingual embedding models. Even for this low resource task, DS-Caps was able to outperform other capsule network-based methods, as well as the recurrent models.

\section{Related Work}\label{related}



Rather recent experiments suggest approaches based on combinations of sequential models like RNNs, LSTMs, and CNNs to capture both contextual information and n-gram features of text~\cite{wang2016combination}. Yet, these approaches still suffer from inherent limitations of sequential text processing with RNNs based approaches and convolutions over sequences of text.~\citet{zhang2018sentence} argue that, in order to encode deep language representations within a neural language model, both n-gram features, as well as sequential information of text, should be utilized within a single language model. Their language modeling procedure includes S-LSTM, a graph RNN. S-LSTM  encodes each word of a sentence as a separate state, as well as the sentence as a global state in each recurrent step. For this, they utilize message passing over graphs~\cite{scarselli2008graph} in order to capture the deep neural representation of languages.

Furthermore, capsule networks that were initially employed in the image processing tasks produced state-of-the-art results with the dynamic routing procedure proposed by~\citet{sabour2017dynamic}. The intention behind the capsule strategy was to represent the features of objects within the data as vector representation in order to identify the exact order or pose of the information. The dynamic routing procedure between capsules mitigates the issues of information loss of CNNs due to max-pooling and elevates the advancement of the part-to-whole relationship between capsules for deeper capsule representation for image classification tasks.

Inspired by the impressive performance of the capsule network architectures for image classification,~\citet{wang2018sentiment} applied the same for sentiment classifications with the combination of RNNs, which produced state-of-the-art results at that time.~\citet{zhao2018investigating} conducted an empirical experiment of capsule networks with dynamic routing to validate utilization of capsule networks for text classification. The implementation of capsule-A and capsule-B, considering different n-gram variations produced the optimal performances for text classification tasks. With even deeper analysis,~\citet{kim2020text} produced an approach based on static routing between capsules for text classification tasks. This procedure alleviates the limitations provided by the variations of text with background noise for capsules with dynamic routing. However, capsules that are solely built upon convolutions do not illustrate the ability to capture the context of the text when the length of sequences increases. 


\section{Model Architecture}

\begin{figure}[h]
  \centering
  \includegraphics[scale=0.27]{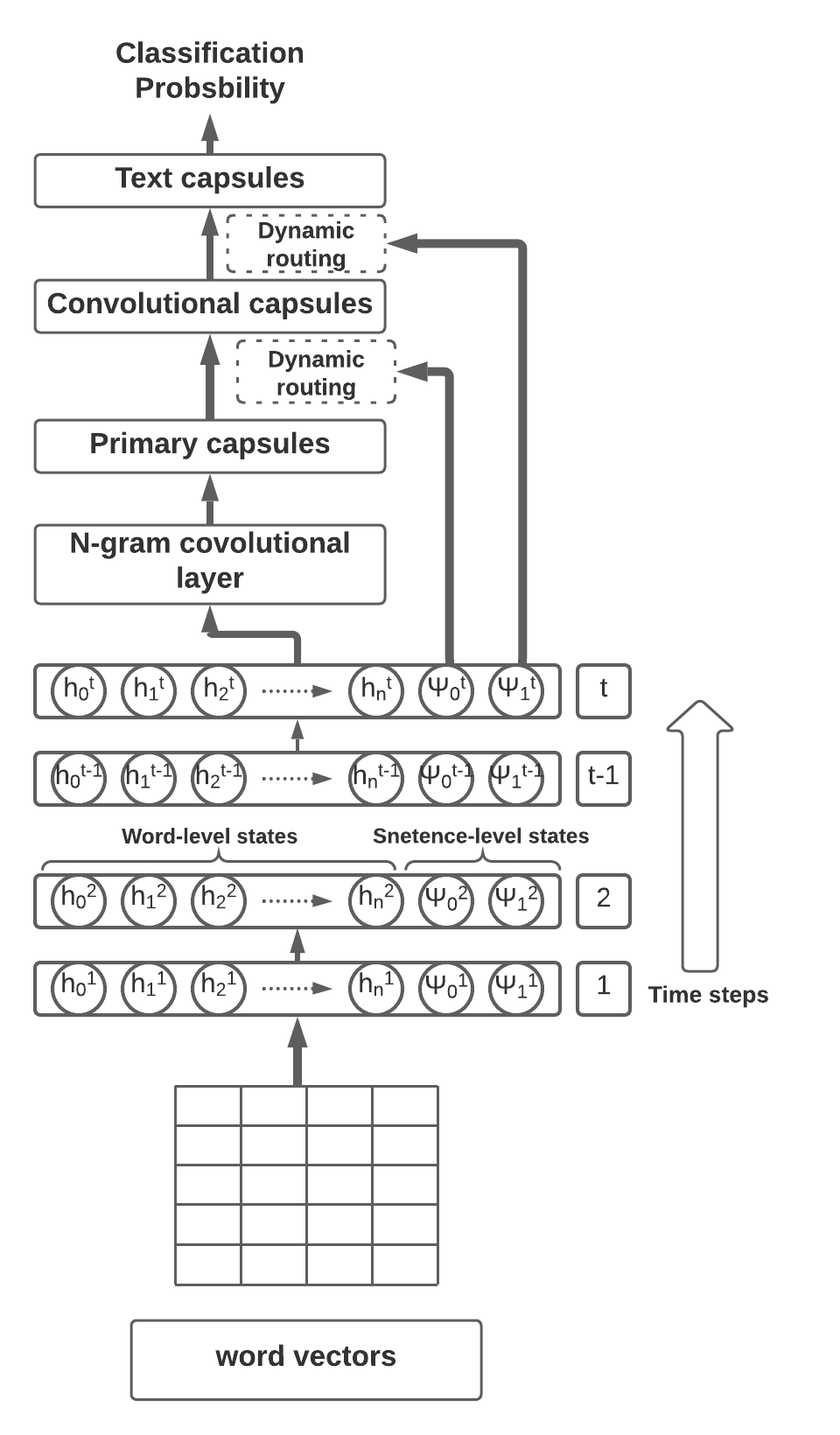}
  \caption{Dual-State Capsule Networks}
  \label{fig:capsnet}
\end{figure}

DS-Caps further improves the language representation capability of capsule networks~\cite{zhao2018investigating}  with the use of sentence-level states and word-level states, an idea borrowed from the S-LSTM architecture~\cite{zhang2018sentence}. As illustrated in Figure~\ref{fig:capsnet}, for the purpose of capturing n-grams features with sequential dependencies of the text, sentence-level states, and word-level states were employed on top of feature embeddings. 

In particular, vanilla capsule models leverage pre-trained word-embeddings as the input features. The idea behind the word-level states is to capture both n-gram and sequential features of the text simultaneously and corporate with vector representation of capsules to enhance language modeling. Therefore instead of using word-embeddings as feature inputs for vanilla capsule, feature maps based on concatenated word-level states were utilized.

Furthermore, the sentence-level states which carry global context-level information were fed into the dynamic routing process between capsules. This improves the context-sensitivity of capsules compared to vanilla capsule networks considering the variability of background information of text.

\subsection{Sentence-level and Word-level States}\label{slstm-discussion}

For the purpose of extracting n-gram features while preserving sequential dependencies of text, \textbf{(1.) global sentence-level states were utilized for context-level feature extraction, and (2.) word-level states were employed for each word to extract dependencies among words in the sequences of text as n-gram features}.

Generally, an encoded hidden state of a sentence using sentence-level states ($\Psi_1^t$ and $\Psi_2^t$) and word-level states ($h_i^t$) at a given time step $t$ could be represented as Eq.~\ref{eq:general}.
\begin{equation}\label{eq:general}
H^t ={\langle}h_1^t,h_2^t,h_3^t,......,h_{n}^t,\Psi_1^t,\Psi_2^t{\rangle}.
\end{equation}

Here the state of the neural representation of encoded hidden state $H^t$ of S-LSTM layer consists of a sub-state $h_i^t$ for each word $w_i$. Since two dynamic routing procedures exist between three capsule layers in a vanilla capsule network, two sentence-level sub-states $\Psi_1^t$ and $\Psi_2^t$ are introduced to utilize towards the optimizations of dynamic routing procedure.

The suggested approach for the neural encoding of sentences utilizes recurrent information exchange between word-level states and sentence-level states to incrementally achieve a rich neural representation in each time step. The initial states are set as $H^0$ = $h_i^0$ = $\Psi_1^0$ = $\Psi_1^0$ = $h^0$ in the form of model parameters. The state transition of a word state $h_i^t$ could be illustrated with the following equations.

\begin{equation}\nonumber
\begin{gathered}
{\phi}_i^t = [h_{i-1}^{t-1},h_i^{t-1},h_{i+1}^{t-1}]\\
\Tilde{\Psi} = avg(\Psi_{1}^{t-1},\Psi_{2}^{t-1})\\
\widehat{i}_i^t = \sigma(W_i\phi_i^t+U_{i}x_i+V_{i}\Tilde{\Psi}+b_i)\\
\widehat{l}_i^t = \sigma(W_l\phi_i^t+U_{l}x_i+V_{l}\Tilde{\Psi}+b_l)\\
\widehat{r_i^t} = \sigma(W_r\phi_i^t+U_{r}x_i+V_{r}\Tilde{\Psi}+b_r)\\
\widehat{f}_i^t = \sigma(W_f\phi_i^t+U_{f}x_i+V_{f}\Tilde{\Psi}+b_f\\
\widehat{s}_{i1}^t = \sigma(W_{s1}\phi_i^t+U_{s1}x_i+V_{s1}\Tilde{\Psi}+b_{s1})\\
\widehat{s}_{i2}^t = \sigma(W_{s2}\phi_i^t+U_{s2}x_i+V_{s2}\Tilde{\Psi}+b_{s2})\\
{o_i^t} = \sigma(W_o\phi_i^t+U_{o}x_i+V_{o}\Tilde{\Psi}+b_o)\\
{u_i^t} = {tanh}(W_u\phi_i^t+U_{u}x_i+V_{u}\Tilde{\Psi}+b_u)\\
i_i^t,l_i^t,r_i^t,f_i^t,s_i^t = {softmax}(\widehat{i}_i^t,\widehat{l}_i^t,\widehat{r}_i^t,\widehat{f}_i^t,\widehat{s}_{i1}^t,\widehat{s}_{i2}^t)\\
\begin{split}
c_i^t = l_i^t{\odot}c_{i-1}^{t-1}+f_i^t{\odot}c_i^{t-1}+ & r_i^t{\odot} c_{i+1}^{t-1}+s_{i1}^t{\odot}c_{\Psi_1}^{t-1}\\  
  & +s_{i2}^t{\odot}c_{\Psi_2}^{t-1}+i_i^t{\odot}u_i^t
\end{split}\\
h_i^t = o_i^t{\odot}{tanh(c_i^t)}\\
\end{gathered}
\label{eq:10}
\end{equation}

Here, {$\phi_i^t$} is the representation of concatenated hidden vectors of the context window. {$\Tilde{\Psi}$} represents the averaged state of two sentence-level states $\Psi_1$ and $\Psi_2$. Following the S-LSTM network~\cite{zhang2018sentence} six gates were applied to control information flow from cell state of previous time step, left cell state of previous time step, right cell state of previous time step, two global sentence states of previous time step, and input state of a given word to cell state of any given time step. These gates are denoted respectively as $f_i^t$, $l_i^t$, $r_i^t$, $s_{i1}^t$, $s_{i2}^t$ and $i_i^t$. $o_i^t$ represents the gate that controls the information flow from the current cell state $c_i^t$ to the current hidden state $h_i^t$. $W_x$, $U_x$, $V_x$ and $b_x$ are model parameters where x $\in$ \{i, o, l, r, f, s1, s2, u\}.

The following equations illustrate the state transitions for global sentence-level states $\Psi_1$ and $\Psi_2$.
\begin{equation}\nonumber
\begin{gathered}
\Tilde{h} = avg(h_1^{t-1},h_2^{t-1},...,h_{n}^{t-1})\\
\widehat{f}_{\Psi_1}^t = \sigma(W_{\Psi_1}\Psi_1^{t-1}+U_{\Psi_1}\Tilde{h}+b_{\Psi_1})\\
\widehat{f}_{\Psi_2}^t = \sigma(W_{\Psi_2}\Psi_2^{t-1}+U_{\Psi_2}\Tilde{h}+b_{\Psi_2})\\
\widehat{f}_{i}^t = \sigma(W_{f}\Tilde{\Psi}+U_{fi}h_i^{t-1}+b_{fi})\\
{o_{\Psi_1}^t} = \sigma(W_{o1}\Psi_1^{t-1}+U_{o1}\Tilde{h}+b_{o1})\\
{o_{\Psi_2}^t} = \sigma(W_{o2}\Psi_2^{t-1}+U_{o2}\Tilde{h}+b_{o2})\\
\begin{split}
f_1^t,f_2^t,..,f_{n}^t,{f_{\Psi_2}^t},{f_{\Psi_2}^t} = {softmax}(\widehat{f}_1^t,
& \widehat{f}_2^t,..,\widehat{f}_{n}^t,\\
\widehat{f}_{\Psi_1}^t,\widehat{f}_{\Psi_2}^t)
\end{split}\\ 
c_{\Psi_1}^t = f_{\Psi_1}^t{\odot}c_{\Psi_1}^{t-1} + \sum{f_{i1}^t{\odot}c_i^{t-1}}\\
c_{\Psi_2}^t = f_{\Psi_2}^t{\odot}c_{\Psi_2}^{t-1} + \sum{f_{i}^t{\odot}c_i^{t-1}}\\
\Psi_1^t = o_1^t{\odot}{tanh}(c_{\Psi_1}^t)\\
\Psi_2^t = o_2^t{\odot}{tanh}(c_{\Psi_2}^t)\\
\end{gathered}
\end{equation}

Here, $f_1^t$, $f_2^t$, ..., $f_{n}^t$ and $f_{\Psi_1}^t$, $f_{\Psi_2}^t$ are gates controlling information respectively from $c_1^{t-1}$, $c_2^{t-1}$, ......, $c_{n}^{t-1}$ and $c_{\Psi_1}^{t-1}$, $c_{\Psi_2}^{t-1}$ to $c_{\Psi_1}^t$ and $c_{\Psi_2}^{t}$. $o_{\Psi_1}^t$ and $o_{\Psi_2}^t$ are gates that manipulate information flow from $c_{\Psi_1}^t$ and $c_{\Psi_2}^t$ to ${\Psi_1}$ and ${\Psi_2}$. $W_x$, $U_x$ and $b_x$ are model parameters where x $\in$ \{${\Psi_1}$,${\Psi_2}$, f, o1, o2\}.


\subsection{Capsule Networks}

As displayed in Figure~\ref{fig:capsnet}, As a major enhancement to vanilla capsule network consisting of 4 layers namely n-gram convolutional layer, primary capsules, convolutional capsules and text capsules, concatenated word-level states, ${\langle}h_1^t,h_2^t,h_3^t,......,h_{n}^t{\rangle}$ were fed as input features. Furthermore, two global sentence-level states ($\Psi_{1}$ and $\Psi_{2}$) that are consolidated with context-level-information, are integrated with the dynamic routing process between primary capsules and convolutional capsules, and between convolutional capsules and text capsules~\cite{zhao2018investigating,kim2020text}. We empirically evaluate the usage of the number of sentence-level states for optimal performance in Section~\ref{ablation_experiment}.

\subsubsection{N-gram Convolutional Layer}

For the purpose of extracting n-gram features from different positions of the concatenated word-level states ($h_i^t$), a standard convolutional layer was applied. Let $H \in {\rm I\!R}^ {d \times l}$ denotes the concatenated word-level states with $l$ number of $d$ dimensional word-level states. The convolution operation consists of filters of $C \in {\rm I\!R}^{d \times f}$, which are applied to a context window of size $f$ on top of the concatenated word states $H^{\{i\colon i+f\}}$, where $i$ is the starting index of the context window. This produces new features governed by the Eq.~\ref{eq:a1}.

\begin{equation}\label{eq:a1}
x_{i} = \sigma(H^{\{i\colon i+f\}} \circ C + b_0)
\end{equation}

Here, $x_{i} \in {\rm I\!R}$ represents the generated feature, and $\circ$ denotes the element-wise multiplication between context window and filter while $\sigma$ and $b_0$ are the non-linear activation function (ReLu or tanh) and bias term respectively. This whole convolution process generates $ M_i \in {\rm I\!R}^{d - {f} +1}$ as feature column consists of features obtained through Eq.~\ref{eq:a1}.With N number of filters, ultimately N number of feature columns were generated according to the following Eq.~\ref{a5}.




\begin{equation}\label{a5}
M = [M_1,M_2, ...,M_N] \in {\rm I\!R}^{(d-f+1) \times N}
\end{equation}



\subsubsection{Primary Capsules}

To convert the scalar output of the convolutional layer to the vector representation of capsules, a capsule layer was applied with $v_i \in {\rm I\!R}^u$  as the instantiated parameters of the capsules with dimension $u$. The capsules are generated based on the matrix multiplication that includes matrix filters of $F \in {\rm I\!R}^{N \times u}$. $(d-f+1)$ number of capsules were generated with the matrix multiplication process governed by the following Eq.~\ref{a3}.

\begin{equation}\label{a3}
v_i = squash(F \otimes M_{i} + b_1)
\end{equation}

Here, $v_i$ consists of a map of capsules that includes $(d-f+1)$ number of capsules. $\otimes$ and $b_1$ denote the matrix multiplication operation and bias term, respectively. The squash function refers to the non-linearity displayed in Eq~\ref{squash}. With $D$ number of filters, $D$ number of maps were created, which consist of $d$ dimensional $(d-f+1)$ number of capsules. Therefore, the primary capsule layer could be represented as $V \in {\rm I\!R}^{(d-f+1) \times D \times u}$.





\subsubsection{Convolutional Capsules}

Inspired by the convolutional capsules introduced by~\citet{zhao2018investigating}, a convolutional capsule layer was applied immediately after the primary capsule layer, where a local region from the primary capsule layer is connected to the capsules of the convolutional capsule layer. This procedure improves the model performance due to the nature of the text, where certain objects of text within the local regions of text could get activated when identifying the most probable features towards the final outcome of the classification process.\\
Let the primary capsules in the region $l \times D$ to be mapped to the convolutional capsules using the weight matrix $W^c \in  {\rm I\!R}^{E\times u\times u}$, where $l \times D$ number of capsules are connected to each convolutional capsule to incrementally learn child-to-parent relationship. Here $E$ is the number of parent capsules in the convolutional layer. Given each child capsule, the parent convolutional capsules are generated according to the following Eq.~\ref{pred-vectors}.

\begin{equation}\label{pred-vectors}
{\widehat{u}_{j|i}} = W_j^{c}u_i + {\widehat{b}_{j|i}}
\end{equation}


Here ${\widehat{b}_{j|i}}$ is the bias term for the mapping. The child capsules in the region $l \times D$ are denoted as $u_i$, which are mapped to the parent convolutional capsules. $W^{c}_j$  is the $j^{th}$ weight matrix for the learning by agreement procedure, which maps $u_i$ to the ${\widehat{u}_{j|i}}$. Thus, $(d-f-l+2) \times D$ total number of $u$ dimensional capsules were generated as convolutional capsules.

\subsubsection{Text Capsules}

The final capsule layer was designed to contain a number of capsules based on the number of classes in the text classification tasks. The capsules in the layer below were transformed based on the matrix multiplication process. This generates text capsules while learning the child-to-parent relationship process through the routing procedure.\\
The convolutional capsules were flattened to a list of capsules and were transformed into text capsules by a transformation matrix $W^{d}\in {\rm I\!R}^{Z\times d\times d}$  considering the routing procedure to learn child-to-parent relationship. The final capsules were produced as $x_j \in {\rm I\!R}^d$. Activation of the capsule denotes the class probability of a given text category as $a_j \in {\rm I\!R}$ for each class. Here, $Z$ denotes the number of text capsules.

\subsubsection{Dynamic Routing Algorithm}
The routing by agreement procedure under the dynamic routing algorithm incrementally ensures that appropriate child capsules are sent to parent capsules by iteratively creating a non-linear mapping between capsules~\cite{sabour2017dynamic}. Under this experiment, two dynamic routing procedure exist in the proposed architecture, between primary capsules and convolutional capsules, and between convolutional capsules and text capsules. 

The dynamic routing procedure is initiated by initializing the log prior probabilities, which are the core components for the routing by agreement. As the proposed enhancement, \textbf{instead of initializing the log prior probabilities to values in statistical distribution or zeros, the log prior probabilities are assigned with values obtained through reshape operation on, global sentence-level states.} This strategy carries context-level information of the text to validate the routing procedure which incrementally improves child-to-parent relationship understanding background knowledge. The log prior probabilities are initialized as in Eq.~\ref{coupling_coefficient}.

\begin{equation}\label{coupling_coefficient}
b_{ij} \gets reshape(\Psi)
\end{equation}

Here $b_{ij}$ represents the log prior probability between the child capsule $i$ and the parent capsule $j$, while the reshape operation includes a matrix multiplication that maps the dimensions of sentence-level state $\Psi$ to the dimensions of $b_{ij}$ ( Both $\Psi_{1}$ and $\Psi_{2}$ sentence-level states were considered under this operation.) The log-le prior probabilities between each child capsule $i$ and each parent capsule $j$ were normalized using the standard $softmax$ function during the iterative routing procedure, which sums up coupling coefficients of each child capsule $i$ to all parent capsules $j$ in the layer above, to 1 as the Eq.~\ref{softmax}.

\begin{equation}\label{softmax}
c_{ij} \gets \frac{exp(b_{ij})}{\sum_{k}{exp(b_{ik})}}
\end{equation}

Here, $c_{ij}$ represents the coupling coefficient that is used when each child capsule $i$ in the layer below maps to each parent capsule $j$ in the layer above. These log prior probabilities $b_{ij}$ could be iteratively learned at the same time as all other weights, where the log prior probabilities only depend on the location of the child and parent capsules but not on the given sequence of the input text. Then the log prior probabilities between the child and parent capsules are updated in an iterative manner, considering the agreement measurement that repeatedly measures the similarity among predicted parent vectors by child capsules $\widehat{u}_{j|i}$ ( predicted parent capsule $j$ given child capsule $i$ ) and current capsule $s_j$. The similarity measurement is a simple scalar product between the predicted and current parent capsules.

\begin{equation}\label{cueernt_parent}
s_{j} = \sum_i {c_{ij}*{\widehat{u}_{j|i}}}
\end{equation}

\begin{equation}\label{updateing_prob}
b_{ij} \gets b_{ij} + \widehat{u}_{j|i} * s_j
\end{equation}

The current parent capsule is calculated based on the coupling coefficients between the child and parent capsules ${c_{ij}}$, and the predicted parent capsules by the child capsules denoted as $\widehat{u}_{j|i}$ according to the Eq.~\ref{cueernt_parent}. The log prior probabilities $b_{ij}$ are iteratively updated according to the Eq.~\ref{updateing_prob} considering the measurement of the agreement between current and predicted parent capsules.

After the iterative routing procedure, the output capsules are normalized using the non-linear squash function, which transforms the length of each vector to represent the probability of the existence of the entity present within a capsule. The squash function converts the length of long vectors near to 1 while shrinking the length of short vectors to 0, which represents the probability as the potential of an entity represented by each capsule. The non-linear squash function applied for parent capsules $s_{j}$  that outputs $v_{j}$ is defined as Eq.~\ref{squash}.

\begin{equation}\label{squash}
v_{j} = \frac{\|s_j\|^2}{1+\|s_j\|^2}\frac{s_j}{\|s_j\|^2}
\end{equation}

Here $\|s_j\|$ denotes the standard norm for $s_j$, and as illustrated in \textbf{Algorithm 1}, the non-linear squash function is applied outside the iterative mapping procedure. This results in eliminating the degradation of log prior probabilities in each routing iteration.


\begin{algorithm}\label{dymaic-algo}

\DontPrintSemicolon
\textbf{procedure} ROUTING({\^u}$_{j|i}$,$\Psi$, $r$, $l$ )\;{
        \ForAll{capsules $i$ in $layer\;l$ and $j$ in $layer\;{l+1}$}   
        { 
        $b_{ij}\gets reshape(\Psi)$ \;
        }
        \For{$r$ iterations}{
        \ForAll{capsules i in $layer \;l$}{
        $c_{i} \gets $$softmax(b_{i})$$ $
        }
        \ForAll{capsules j in $layer\;{l+1}$}{
        $s_j \gets $$\sum_{i} c_{ij}*\widehat{u}_{ij}$$ $
        }
        \ForAll{capsules $i$ in $layer\;l$ and $j$ in $layer\;{l+1}$}{
        $b_{ij}\gets b_{ij}+u_{j}*s_j$
        }
        }
        
        \ForAll{capsules $j$ in $layer_{l+1}$ }
        {
        $v_j \gets squash(s_j)$ \;
        }
\textbf{return} $v_j$ \;
}
\caption{Dynamic Routing using Sentence State}
\end{algorithm}

\subsubsection{Loss Function}

For the text classification task, for each text capsule, we used a separate margin loss~\cite{sabour2017dynamic} function to identify where a given text category is present within a given capsule. For text capsule $s$, the margin loss $L_s$ is given by;


\begin{multline}
L_{s} = T_s{max(0,m^{+}-{\|v_s\|})^2} \\
+ \lambda{(1-T_s)}{max(0,{\|v_s\|- m^-})^2}
\end{multline}
Here $T_s = 1$ if the text category exists within the text capsule, otherwise it is set to 0. $m^+$ and $m^-$ are set as 0.9 and 0.1 accordingly. The down-weighting coefficient $\lambda$ is set to 0.25 with the optimal performance.


\section{Experiments}
\subsection{Data Sets}

Experiments were conducted on six benchmark datasets covering multiple classification tasks. These include movie review classification ( MR(2004)~\cite{pang2004sentimental}, MR(2005), IMDB ), news article classification ( Reuters10, MPQA~\cite{wiebe2005annotating}  ), and question categorization ( TREC-QA~\cite{li2002learning} ). The details of each dataset are shown in table \ref{tab : data-sets}. For measuring the performance of the model against low resource language processing, a publicly available Sinhala dataset~\cite{Senevirathne2020deep} was utilized. This includes 15059 news comments annotated with 4 sentiment categories with an average comment length of 10.







\subsection{Implementation}

For the experimental analysis, we utilized the 300-dimensional GloVe\footnote{\url{https://nlp.stanford.edu/projects/glove/}} word vectors, which consisted of 840 billion words. Adam optimizer was used for the optimization process with exponential learning rate decay. The models were trained on Google Colab with Tensorflow as the implementation tool.  The optimal hyperparameters for models in each data set are indicated in Table~\ref{tab : hyper-parameters}.  


For each experiment, the learning rate was set to $1e-3$, and the learning rate decay was set to 0.95. The hidden word-level state dimension was set to 300, while sentence-level hidden state size was set to 600 dimensions. Max sentence length was chosen as the sentence length, considering the variations of the datasets. The n-gram convolutional layers were instantiated to extract 3-grams from each context window with 32 filters. Each capsule in the primary capsule layer was instantiated with 8-dimensional vectors, while each convolutional and text capsule was instantiated with 16-dimensional vectors. The length of each capsule denotes the existence of an entity within a capsule, which is further utilized with text capsules to identify the text categories within a given sequence of text.

\begin{table}
\centering
\noindent\adjustbox{max width=\linewidth}{
\begin{tabular}{l||c|c|c|c|c|c}
\hline 
\hline 
\textbf{Data} & \textbf{c} & \textbf{Train} & \textbf{Dev} & \textbf{Test} &  \textbf{$l_{avg}$} & \textbf{$|V|$}\\ 
\hline
\hline
MR(2004)  & 2 & 1620  & 180  & 200  &  779 & 40693 \\
MR(2005)  & 2 & 8635  & 960  & 1067  & 22 & 18764\\
Reuters10  & 10 & 6472  & 720  & 2787  & 168 & 28482 \\
TREC-QA & 6 & 4843  & 539  & 500  & 9 & 8689\\
MPQA & 2 & 8587 & 955  & 1067  & 3 & 6246\\
IMDB  & 2 & 22500  & 2500  & 25000 & 231 & 112540\\
\hline
\end{tabular}
}
\caption{\label{tab : data-sets} Properties of datasets. c: Number of text categories, Train, Dev, Test: Size of training, development, and test sets (respectively), $l_{avg}$: Average sentence length, $|V|$: Size of the vocabulary}
\end{table}

\begin{table}
\centering
\noindent\adjustbox{max width=\linewidth}{
\begin{tabular}{l||c|c|c|c|c}
\hline 
\hline 
\textbf{Data}  & \textbf{Batch} & \textbf{Routing} & \textbf{Steps}  &  \textbf{CW}  &  \textbf{Epochs} \\ 
\hline
\hline 
MR(2004)& 4 & 3 & 4  & 1 & 50  \\
MR(2005)& 8 & 3 & 8  & 2 & 20  \\
Reuters10 & 8 & 2 & 4  & 2 & 20   \\
TREC-QA & 4 & 3 & 4  & 1 & 50 \\
MPQA & 4 & 4 & 4  & 1 & 20  \\
IMDB & 8 & 3 & 8 & 2 & 10 \\

\hline
\end{tabular}
}
\caption{\label{tab : hyper-parameters} Optimal hyperparameters for each dataset. Batch: Batch size, Routing: Number of iterations in the dynamic routing procedure, Steps: Number of recurrent time steps used to encode the sequence of text utilizing word and sentence states, CW: Number of word-states considered in both left and right context for encoding word-states in a particular time step.}

\end{table}

\begin{table*}
\centering
\noindent\adjustbox{max width=\textwidth}{
\begin{tabular}{l|| c|c|c|c|c|c}
\hline 
\hline 
\textbf{}  & \textbf{MR(2004)} & \textbf{MR(2005)} & \textbf{Reuters} & \textbf{TREC-QA} & \textbf{MPQA} & \textbf{IMDB}  \\ 
\hline
\hline

CNN-rand~\cite{kim2014convolutional}   & -  & 76.1  & -  & 91.2 & -  & 83.4\\
CNN-static~\cite{kim2014convolutional}  & -  & 81.0  & -  & 92.8  & - & 89.6 \\
CNN-non-static~\cite{kim2020text} & 88.0  & 81.5  & 87.4  & 92.7  & 89.9  & 90.4\\
\hline
LSTM~\cite{zhao2018investigating}  & -  & 75.9  & -  & 86.8  & -  & - \\
Bi-LSTM~\cite{zhao2018investigating} & -  & 79.3  & -  & 89.6  & -  & -\\
Tree-LSTM~\cite{zhao2018investigating}   & - & 80.7  & -  & 91.8 & -  & - \\
S-LSTM~\cite{zhang2018sentence}   & 82.4  & -  & -  & -  & -  & 87.2 \\
\hline
Capsule-A~\cite{kim2020text}   & 85.0  & 79.4  & 87.7  & 90.8  & 88.3  & 89.3\\
Capsule-B~\cite{kim2020text}    & 89.5  & 79.0  & 88.0  & 89.8  & 88.4  & 89.3\\
Capsule-static routing~\cite{kim2020text}    & 89.6  & 81.0  & 87.5  & \textbf{94.8}  & \textbf{90.6}  & 89.7\\
Dual-state capsule networks (DS-Caps) & \textbf{90.5} & \textbf{82.1}  & \textbf{88.6}  & 92.8 & 89.2  & \textbf{90.6}\\

\hline
\end{tabular}
}
\caption{\label{tab : results} Text classification accuracies for benchmark datasets.}
\end{table*}

\subsection{Baseline models}

In this experiment, we empirically evaluated our models against several baseline models such as LSTM, Bi-LSTM, CNN with randomly initialized vectors, CNN with non-trainable embeddings (CNN-static), CNN with trainable embeddings (CNN-non-static)~\cite{kim2014convolutional} and and S-LSTM~\cite{zhang2018sentence} as primary baseline models. Furthermore, we evaluated the DS-Caps approach against several capsule networks based techniques namely, Capsule-A, Capsule-B~\cite{zhao2018investigating}, and capsule networks with static routing~\cite{kim2020text}. Some evaluation records for datasets under certain baseline techniques are not displayed in Table~\ref{tab : results} as the results are not reported in the literature.

\section{Evaluation}

\subsection{Evaluation on Benchmark Data Sets}
Accuracy was chosen as the evaluation metric for our experiments, following related research in the same domain~\cite{zhao2018investigating}. The experiment results are summarized in the Table~\ref{tab : results}, against the six benchmark datasets. 

Our DS-Caps network was able to achieve the best result for four out of six benchmark results including MR(2004), MR(2005), Reuters and IMDB datasets, outperforming the previous studies on capsule networks with static routing for MR(2004) and MR(2005) datasets, Capsule-B for Reuters10 dataset, and CNN-non-static for IMDB dataset.
In particular, the DS-Caps network extensively and consistently defeats sequential neural networks such as LSTM, BiLSTM, Tree LSTM and S-LSTM networks, and spatial models such as CNN-rand, CNN-static, CNN-non-static on all six datasets. The observation was expected due to the language representation of capsules, where the vector representation could be greatly associated with the exact order or pose of sequences of text. The dynamic routing procedure introduced under the capsule networks further eliminated the loss of information due to the pooling strategy used under CNNs. This observation could be justified as the ability of the S-LSTM layer to encode the neural representation of text more efficiently, which subsequently enriches the dynamic routing procedure providing context-level information through global sentence-level information.\\
Furthermore, the results obtained for datasets with higher sequence lengths (MR(2004) and IMDB) illustrate greater performance compared to the baseline models. This provides shreds of evidence for the capability of the DS-Caps network to classify rather longer sequences of text utilizing sequential features and n-gram features of the text, collaborating with the vector representations of capsules.





\subsection{Evaluation on Low Resource Languages}
As mentioned in Section~\ref{intro}, to demonstrate that DS-Caps gives state-of-the-art results on low resource languages, we selected Sinhala. Sinhala is a low resource language, and the largest reported dataset is the CommonCrawl dataset, which is just above 100M~\cite{lakmal2020word}. Furthermore, Sinhala does not have a pre-trained BERT mode, nor is it included in multiBERT.  Using auxiliary features such as POS tags is also not an option for this language, as the best-reported results for POS tagging are not optimal~\cite{fernando2018evaluation}. \\ 
The proposed DS-caps network was evaluated against a Sinhala multi-class sentiment analysis task~\cite{Senevirathne2020deep}. As the results displayed in Table~\ref{sinhala performance comparison}, DS-caps outperformed all previously tested deep learning approaches for Sinhala sentiment analysis. The hyper-parameters used for the DS-Caps networks were the same as the parameter used by~\citet{Senevirathne2020deep} for Capsule-A and Capsule-B and fastText embeddings were used as primary features for sentiment analysis task.

DS-caps produced promising results in the context of this low resourced language, suggesting the possibility to be effective for any language as a universal approach. Even though BERT models~\cite{devlin2018bert} indicated superior performance in text classification tasks, they cannot be readily applied for low resourced languages due to lack of pre-trained models and high computation bottleneck. Also, the lack of linguistic resources for low resource languages, makes it impossible to use highly resource-intensive methodologies. Therefore DS-caps could be used as a proper substitution in place of text classification models that are based on computationally intensive BERT models.

\begin{table}\label{sinhala_sentiment_results}
\centering
\noindent\adjustbox{max width=\linewidth}{
\begin{tabular}{c|| c|c|c|c}
\hline 
\hline 
\textbf{Model}  & \textbf{Accuracy} & \textbf{Precision} & \textbf{Recall}  &  \textbf{F1-score} \\ 
\hline
\hline
RNN & 58.98 & 42.93 & 54.98  &  42.30  \\
LSTM & 62.88 & \textbf{70.95} & 51.93  & 54.50  \\
GRU & 62.78 & 60.93 & 62.78  &  54.83  \\
Bi-LSTM & 63.81 & 61.17 & 63.81  &  57.71  \\
Stacked-BiLSTM & 63.13 & 69.71 & 63.18  & 59.42  \\
\hline
HAHNN & 61.16 & 71.08 & 48.54  &  59.25  \\
\hline
Capsule-A & 61.89 & 56.12 & 61.89  &  53.55  \\
Capsule-B & 63.23 & 59.84 & 63.23  & 59.11  \\
DS-Caps & \textbf{64.03} & 61.68 &  \textbf{64.03}  & \textbf{61.33} \\

\hline
\end{tabular}
}
\caption{\label{sinhala performance comparison} 10-fold cross-validated, weighted evaluation metrics for performance on Sinhala multi-class sentiment analysis~\cite{Senevirathne2020deep}}
\end{table}

\begin{table}
\centering
\noindent\adjustbox{max width=\linewidth}{
\begin{tabular}{c|c|c|c|c}
\hline  
\hline  
\textbf{No. of Sentence states} & \textbf{Loss function} & \textbf{Context window} & \textbf{Time steps} &  \textbf{Accuracy}  \\ \hline
\hline
0 & Margin  & 2 & 5  & 80.6\\
0  & Cross entropy & 2  & 7  & 79.4\\
0 & Spread   & 2  & 10  & 79.9 \\
0 & Margin + L2  & 3  & 4  & 80.3 \\
\hline
1  & Margin  & 1  & 3  &  80.5 \\
1 & Cross entropy   & 2  & 3  & 79.1 \\
1 & Spread  & 2 & 4  & 80.4\\
1 & Margin + L2 & 2 &  4 & 80.1\\
\hline
2 & Margin    & 1 & 7 & \textbf{82.1}\\
2  & Cross entropy    & 2  & 7 & 79.9 \\
2 & Spread  &  2  & 3 & 79.7\\
2 & Margin + L2 &  2  & 4  & 80.9 \\

\hline
\end{tabular}
}
\caption{\label{tab : ablation components} Performance of DS-Caps networks varying no. of sentence-level states against lost function for MR(2005) dataset.}
\end{table}

\subsection{Ablation Study}\label{ablation_experiment}

For the purpose of analyzing the performance of the model varying different components, an ablation study was conducted on MR(2005) dataset and reported in Table~\ref{tab : ablation components}. The performance was analyzed with respect to loss functions proposed by~\cite{zhao2018investigating} for capsule networks and further adding L2 regularization to the optimal loss function. Furthermore, the effect of the number of sentence-level states was analyzed to find out the optimal number of sentence-level states to be integrated with the dynamic routing process. Under the experiments, the batch size was kept as 8 and, the number of epochs and dynamic routing iterations were fixed as 20 and 3 respectively. Margin loss with two sentence-level states produced the best performance for MR(2005) dataset.

This result was expected due to the efficiency of margin loss for vanilla capsule network~\cite{zhao2018investigating}. Further two distinct sentence-level states facilitate separate learning procedures for dynamic routing algorithms to learn log prior probabilities which elevate part-to-whole relationship considering low-level and high-level-capsules.




\section{Conclusion and Future Work}
In this paper, we proposed a novel Dual-State Capsule (DS-Caps) network architecture, which incrementally improves the language representation considering local and global information of the text. As further enhancements, attention mechanisms could be integrated with DS-Caps. It is also worthwhile to integrate language resources such lexicons and contextual embeddings such as BERT~\cite{devlin2018bert} with DS-Caps since most of the other deep learning approaches have produced greater performances with such strategies~\cite{saha2020bert}.




\bibliography{anthology,eacl2021}
\bibliographystyle{acl_natbib}

\appendix



\end{document}